%% file: main.tex
\documentclass[10pt,twocolumn,letterpaper]{article}
\usepackage{setspace}
\usepackage[pagenumbers]{cvpr} 

\usepackage{multirow}
\usepackage{algorithm}
\usepackage{algorithmicx}
\usepackage{algpseudocode}

\input{preamble}

\definecolor{cvprblue}{rgb}{0.21,0.49,0.74}
\usepackage[pagebackref,breaklinks,colorlinks,citecolor=cvprblue]{hyperref}

\title{PanGu-Draw: Advancing Resource-Efficient Text-to-Image Synthesis with Time-Decoupled Training and Reusable Coop-Diffusion}
\author{
Guansong Lu\textsuperscript{1} \quad Yuanfan Guo\textsuperscript{1} \quad Jianhua Han\textsuperscript{1} \quad Minzhe Niu\textsuperscript{1} \quad Yihan Zeng\textsuperscript{1} \\ 
Songcen Xu\textsuperscript{1} \quad Zeyi Huang\textsuperscript{2} \quad Zhao Zhong\textsuperscript{2} \quad Wei Zhang\textsuperscript{1} \quad Hang Xu\textsuperscript{1} \\
\textsuperscript{1}Huawei Noah’s Ark Lab \quad \textsuperscript{2}Huawei \\
}

\begin{document}
\setstretch{0.98}
\maketitle

\input{arxiv_sec/0_abstract}    
\input{arxiv_sec/1_intro}

\input{arxiv_sec/2_related_work}

\input{arxiv_sec/3_method}
\input{arxiv_sec/4_experiments}

\input{arxiv_sec/5_conclusion}

\clearpage
{
    \small
    \bibliographystyle{ieeenat_fullname}
    \bibliography{main}
}

\input{arxiv_sec/X_suppl}

\end{document}

%% file: preamble.tex
\usepackage[dvipsnames]{xcolor}


%% file: arxiv_sec/0_abstract.tex
\begin{abstract}
Current large-scale diffusion models represent a giant leap forward in conditional image synthesis, capable of interpreting diverse cues like text, human poses, and edges. 
However, their reliance on substantial computational resources and extensive data collection remains a bottleneck. 
On the other hand, the integration of existing diffusion models, each specialized for different controls and operating in unique latent spaces, poses a challenge due to incompatible image resolutions and latent space embedding structures, hindering their joint use.
Addressing these constraints, we present \textbf{``PanGu-Draw"}, a novel latent diffusion model designed for resource-efficient text-to-image synthesis that adeptly accommodates multiple control signals. 
We first propose a resource-efficient Time-Decoupling Training Strategy, which splits the monolithic text-to-image model into structure and texture generators.
Each generator is trained using a regimen that maximizes data utilization and computational efficiency, cutting data preparation by 48\% and reducing training resources by 51\%.
Secondly, we introduce \textbf{``Coop-Diffusion"}, an algorithm that enables the cooperative use of various pre-trained diffusion models with different latent spaces and predefined resolutions within a unified denoising process. This allows for multi-control image synthesis at arbitrary resolutions without the necessity for additional data or retraining.
Empirical validations of Pangu-Draw show its exceptional prowess in text-to-image and multi-control image generation, suggesting a promising direction for future model training efficiencies and generation versatility.
The largest 5B T2I \textbf{PanGu-Draw} model is released on the Ascend platform. Project page: \href{https://pangu-draw.github.io}{https://pangu-draw.github.io}

\end{abstract}

%% file: arxiv_sec/1_intro.tex
\section{Introduction}
\label{sec:intro}

\begin{figure*}[!t]
\vspace{-8mm}
\centering
\includegraphics[width=0.9\linewidth]{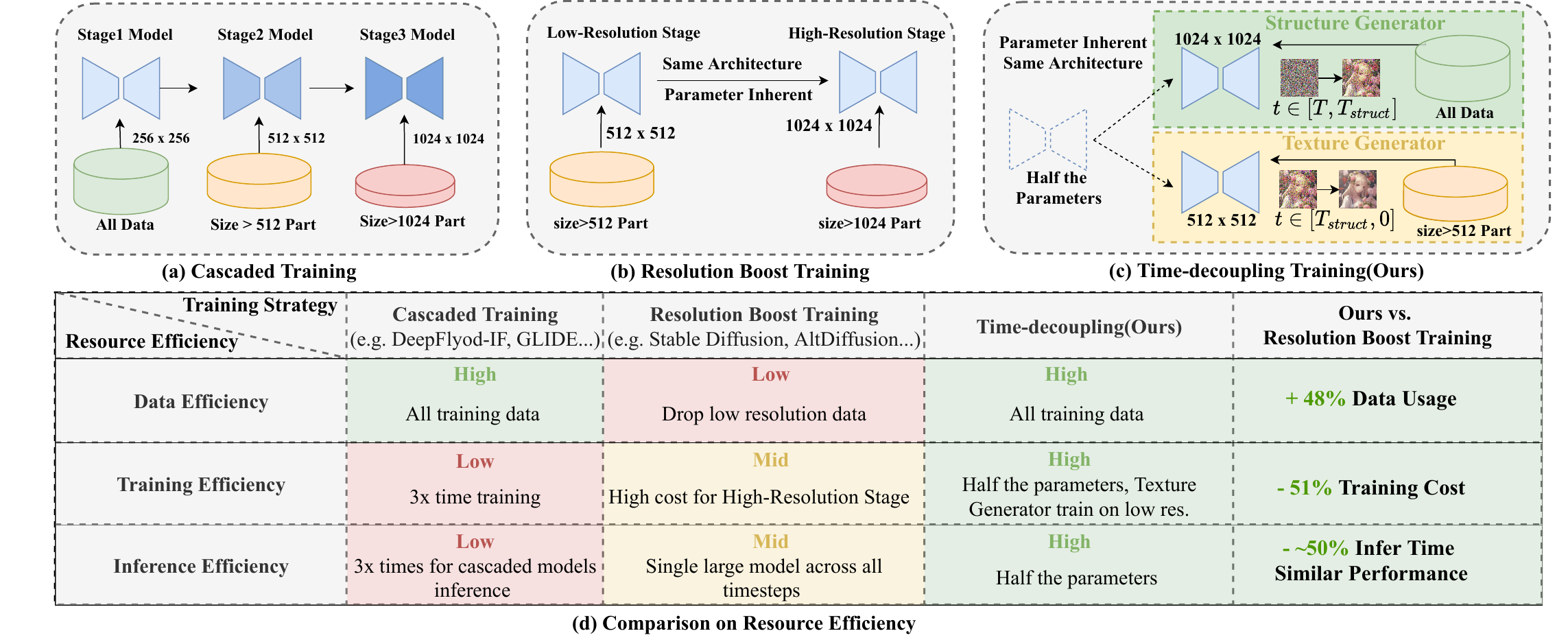}
\caption{Illustration of three multi-stage training strategies and comparison between them in resource efficiency in data, training and inference aspects. Our time-decoupling training strategy significantly surpasses the representative methods in Cascaded Training~\cite{shonenkov2023deepfloyd,nichol2021glide} and Resolution Boost Training~\cite{sd,ye2023altdiffusion} in resource efficiency.}
\label{fig:framework}
\end{figure*}

The Denoising Diffusion Probabilistic Models (DDPMs) \cite{ddpm} and their subsequent enhancements \cite{nichol2021improved-ddpm,adm,ho2022classifier-free} have established diffusion models as a leading approach for image generation. These advancements excel in the application of diffusion models to text-to-image synthesis, yielding high-fidelity results with large-scale models and datasets, supported by substantial computational resources \cite{nichol2021glide,ho2022cascaded-diffusion,imagen,ramesh2022dalle2,sd}. These foundational models, capable of understanding and rendering complex semantics, have paved the way for diverse image generation tasks, accommodating various control signals such as reference images, edges \cite{controlnet}, and poses \cite{controlnet}.

However, the extensive computational demand and significant data collection required by these models pose a substantial challenge. The ambitious goal of higher fidelity and increased resolution in image synthesis pushes the boundaries of model and dataset sizes, escalating computational costs, and environmental impact. Moreover, the aspiration for versatile control and multi-resolution in image generation introduces additional complexity. Existing diffusion models, each tailored for specific controls and operating within distinct latent spaces, face the challenge of integration due to incompatible image resolutions and latent space embeddings, obstructing their concurrent utilization.
This incompatibility not only leads to more resource consumption of retraining but also impedes the joint synthesis of images controlled by multiple factors, thereby limiting the scalability and practical application of such existing generative models. In response to these challenges,  our work introduces a novel paradigm named ``\textbf{PanGu-Draw}" that judiciously conserves training resources while enhancing data efficiency, thereby proposing a resource-efficient pathway forward for diffusion model scalability.

As shown in Figure \ref{fig:framework}, the training strategies of predecessors like DeepFloyd \cite{shonenkov2023deepfloyd} and GLIDE \cite{nichol2021glide}, which employ a cascaded approach, excel in leveraging data across resolutions but suffer from inefficient inference due to their reliance on multiple models. Alternatively, Stable Diffusion \cite{sd} and AltDiffusion \cite{ye2023altdiffusion} use a Resolution Boost Training strategy aiming for cost-effectiveness by refining a single model. However, this strategy falls short on data efficiency.

In light of these considerations, our PanGu-Draw framework advances the field by presenting a Time-Decoupling Training Strategy that segments the training of a comprehensive text-to-image model into two distinct generators: one dedicated to structural outlines and another to textural details. This division not only concentrates on training efforts but also enhances data efficacy. The structural generator is adept at crafting the initial outlines of images, offering flexibility in data quality and enabling training across a spectrum of data calibers; the textural generator, in contrast, is fine-tuned using low-resolution data to infuse these outlines with fine-grained details, ensuring optimal performance even during high-resolution synthesis. This focused approach not only accelerates the training process of our\textbf{ 5B model} but also significantly reduces the reliance on extensive data collection and computational resources, as evidenced by a 48\% reduction in data preparation and a 51\% reduction in resource consumption.

Furthermore, we introduce a pioneering algorithm named \textbf{Coop-Diffusion}, which facilitates the cooperative integration of diverse pre-trained diffusion models. Each model, conditioned on different controls and pre-defined resolutions, contributes to a seamless denoising process. The first algorithmic sub-module addresses inconsistencies in VAE decoders that arise during the denoising process across different latent spaces, ensuring cohesive image quality by effectively reconciling disparate latent space representations. The second sub-module confronts the challenges associated with multi-resolution denoising. Traditional bilinear upsampling for the intermediate noise map, introduced during the denoising process, can undesirably amplify the correlation between pixels. This amplification deviates from the initial Independent and Identically Distributed (IID) assumption, leading to severe artifacts in the final output image. However, our innovative approach circumvents this issue with a single-step sampling method that preserves the integrity of pixel independence, thus preventing the introduction of artifacts. \textbf{Coop-Diffusion} obviates the need for additional data or model retraining, addressing the challenges of multi-control and multi-resolution image generation with scalability and efficiency.

PanGu-Draw excels in text-to-image (T2I) generation, outperforming established models like DALL-E 2 and SDXL, as evidenced by its FID of 7.99 in English T2I. It also leads in Chinese T2I across metrics like FID, IS, and CN-CLIP-score. User feedback highlights a strong preference for PanGu-Draw, aligning well with human visual perceptions. Available on the Ascend platform, PanGu-Draw is efficient and versatile.

In summary, our contributions are manifold:
\begin{itemize}
    \item \textbf{PanGu-Draw}: A resource-efficient diffusion model with a Time-Decoupling Training Strategy, reducing data and training resources for text-to-image synthesis.
    \item \textbf{Coop-Diffusion}: A novel approach for integrating multiple diffusion models, enabling efficient multi-control image synthesis at multi-resolutions within a unified denoising process.
    \item Comprehensive evaluations demonstrate \textbf{PanGu-Draw} (5B model) can produce high-quality images aligned with text and various controls, advancing the scalability and flexibility of diffusion-based image generation.
\end{itemize}

%% file: arxiv_sec/2_related_work.tex
\vspace{-1mm}
\section{Related Work}
\vspace{-1mm}
\label{sec:related_work}

\looseness=-1
\noindent\textbf{Text-to-Image Generation.} The integration of diffusion models into the realm of text-to-image generation marks a significant stride in computational creativity \cite{ho2020ddpm,nichol2021improved-ddpm,adm,nichol2021glide,ho2022cascaded-diffusion,ramesh2022dalle2,imagen,sd,xu2022versatile,shonenkov2023deepfloyd,feng2023ernie,xue2023raphael}. 
Text-to-image synthesis models like GLIDE \cite{nichol2021glide} and DALL-E 2 \cite{ramesh2022dalle2}, which incorporate CLIP image embeddings, have significantly advanced in generating diverse and semantically aligned images from textual descriptions. The Latent Diffusion model \cite{sd} addresses computational challenges by creating images from text-conditioned low-dimensional latent representations. Techniques like LoRA \cite{hu2021lora} enhance domain-specific adaptability through low-rank matrix-driven parameter offsets, avoiding catastrophic forgetting. Additionally, ControlNet \cite{controlnet} introduces spatial conditioning controls, offering flexibility in image generation under varied conditions like edges and depth. Current research also focuses on aligning model outputs with human aesthetic preferences, aiming to optimize image quality and user satisfaction \cite{hao2022optimizing,lee2023aligning,wu2023better,xu2023imagereward,dong2023raft}.
Despite the proliferation of such specialized models, a unified framework that consolidates these disparate capabilities remains absent, limiting the potential for multi-control and complex editing in image synthesis.

\noindent\textbf{Model Efficient Training and Scaling Up Strategies.} 
Efficient training and scaling of models are pivotal for advancing large-scale neural networks. In the realm of text-to-image (T2I) diffusion models, the quest for efficiency has led to innovative training strategies. Historical methods, such as those utilized by DeepFloyd \cite{shonenkov2023deepfloyd} and GLIDE \cite{nichol2021glide}, capitalize on cascaded approaches that proficiently utilize data across various resolutions, yet their reliance on multiple models results in less efficient inference processes. Contrastingly, models like Stable Diffusion \cite{sd} and AltDiffusion \cite{ye2023altdiffusion} adopt Resolution Boost Training strategies that refine a single model for cost-effectiveness. Despite the advantages, such strategies do not fully exploit data efficiency. In scaling up strategies, training efficiency is also important. The correlation between model size and performance is well-documented \cite{kaplan2020scaling,hoffmann2022training}, with larger models like SDXL \cite{podell2023sdxl} showing notable gains. Efficient adaptation and scaling are explored in \cite{chen2021bert2bert} through distillation, and in \cite{qin2022elle} by marrying model expansion with domain-specific prompts. Serial scaling and knowledge distillation reduce training times significantly as demonstrated by \cite{fu2023triple}, while \cite{ding2023network} proposes progressive network expansion for faster training with minimal loss. Our approach offers a novel approach to diffusion model scaling that enhances efficiency.

%% file: arxiv_sec/3_method.tex
\vspace{-2mm}
\section{Preliminary}
\vspace{-1mm}
Given an image $x_0$, diffusion models first produce a series of noisy images $x_1,...,x_T$ by adding Gaussian noise to $x_0$ according to some noise schedule given by $\bar{\alpha}_t$ as follows:
\begin{align}
\label{eq:q_sample}
    x_t = \sqrt{\bar{\alpha}_t}x_{0} + \sqrt{1-\bar{\alpha}_t} \epsilon,
\end{align}
where $\epsilon \sim \mathcal{N}\left(0, I\right)$.

Diffusion models then learn
a denoising model $\epsilon_\theta(x_t, t)$ to predict the added noise of a noisy image $x_t$ with the following training objective:
\begin{equation}
    \mathcal{L} = \mathbb{E}_{x_0 \sim q\left(x_0\right), \epsilon \sim \mathcal{N}(0, I), t \sim[1,T]}\left\|\epsilon-\epsilon_\theta\left(x_t, t\right)\right\|^2,
\end{equation}
where $t$ is uniformly sampled from $\{1,...,T\}$. Once the denoising model $\epsilon_\theta(x_t, t)$ is learned, starting from a random noise $x_T \sim \mathcal{N}(0, I)$, one can iteratively predict and reduce the noise in $x_t$ to get a real image $x_0$.
During the sampling process, we can predict the clean data $x_0$ from $\epsilon_\theta\left(x_t, t\right)$ with single-step sampling as follows:
\begin{equation}
\label{eq:epsilon_to_x0}
    \hat{x}_{0,t} = \frac{1}{\sqrt{\bar{\alpha}_t}} (x_t - \sqrt{1 - \bar{\alpha}_t} \epsilon_{\theta} (x_t, t)).
\end{equation}

Our text-to-image generation model is built on the model architecture proposed in Latent Diffusion Model \cite{sd}. In this model, a real image $x_0$ is first down-sampled 8 times as a lower-dimension latent code $z_0$ with an image encoder model $E$, which can be decoded with a latent decoder model $D$ back to a real image $x_0$. The denoising network $\epsilon_\theta(z_t, t, c)$ is parameterized as a U-Net \cite{ronneberger2015u} model, where embedding of time step $t$ is injected with adaptive normalization layers and embedding of input text $c$ is injected with cross-attention layers. 

\begin{figure*}[!t]
\vspace{-8mm}
\centering
\includegraphics[width=0.9\linewidth]{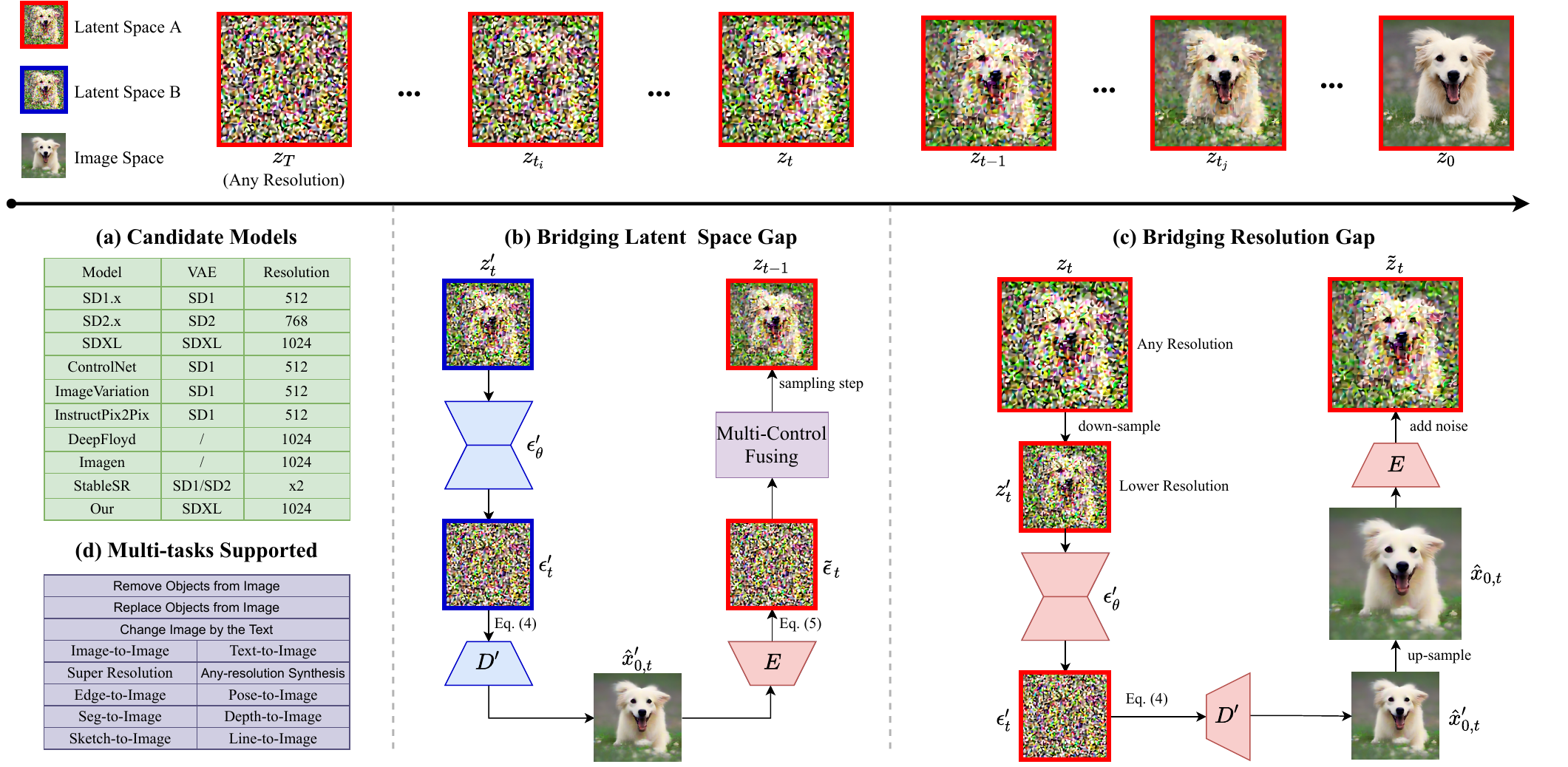}
\vspace{-3mm}
\caption{Visualization of our \textit{Coop-Diffusion} algorithm for the cooperative integration of diverse pre-trained diffusion models. 
(a) Existing pre-trained diffusion models, each tailored for specific controls and operating within distinct latent spaces and image resolutions.
(b) This sub-module bridges the gap arising from different latent spaces by transforming $\epsilon_t'$ in latent space B to the target latent space A as $\tilde{\epsilon}_t$. (c) This sub-module bridges the gap arising from different resolutions by performing upsampling on the predicted clean data $\hat{x}_{0,t}'$.
 }
\label{fig:coop-diffuse}
\vspace{-3mm}
\end{figure*}

\vspace{-1mm}
\section{PanGu-Draw}
\vspace{-1mm}
In this section, we first illustrate our resource-efficient 5B text-to-image generation model, trained with a time-decoupling training strategy and further enhanced with a prompt enhancement LLM.
Then, we present our \textit{Coop-Diffusion} algorithm for the cooperative integration of diverse pre-trained diffusion models, enabling multi-control and multi-resolution image generation.

\subsection{Time-Decoupling Training Strategy}

Enhancing data, training, and inference efficiency is vital for text-to-image models' practical use. Figure \ref{fig:framework} shows two existing training strategies: (a) Cascaded Training, using three models to incrementally improve resolution, is data-efficient but triples training and inference time. (b) Resolution Boost Training starts at 512x512 and then 1024x1024 resolution, discarding lower resolution data and offering moderate efficiency with higher training costs and single-model inference across all timesteps. These approaches differ from our time-decoupling strategy, detailed below.

Responding to the need for enhanced efficiencies, we draw inspiration from the denoising trajectory of diffusion processes, where initial denoising stages primarily shape the image's structural foundation, and later stages refine its textural complexity. With this insight, we introduce the Time-Decoupling Training Strategy. This approach divides a comprehensive text-to-image model, denoted as $\epsilon_{\theta}$, into two specialized sub-models operating across different temporal intervals: a structure generator, $\epsilon_{struct}$, and a texture generator, $\epsilon_{texture}$. Each sub-model is half the size of the original, thus enhancing manageability and reducing computational load.

As illustrated in Figure \ref{fig:framework}(c), the structure generator, $\epsilon_{struct}$, is responsible for early-stage denoising across larger time steps, specifically within the range ${T, ..., T_{struct}}$, where $0 < T_{struct} < T$. This stage focuses on establishing the foundational outlines of the image. Conversely, the texture generator, $\epsilon_{texture}$, operates during the latter, smaller time steps, denoted by ${T_{struct}, ..., 0}$, to elaborate on the textural details. Each generator is trained in isolation, which not only alleviates the need for high-memory computation devices but also avoids the complexities associated with model sharding and its accompanying inter-machine communication overhead.

In the inference phase, $\epsilon_{struct}$ initially constructs a base structural image, $z_{T_{struct}}$, from an initial random noise vector, $z_T$. Subsequently, $\epsilon_{texture}$ refines this base to enhance textural details, culminating in the final output, $z_0$. This sequential processing facilitates a more resource-efficient workflow, significantly reducing the hardware footprint and expediting the generation process without compromising the model's performance or output quality, as demonstrated in our ablated experiment in Sec. \ref{sec:ablation_study}.

\noindent \textbf{Resource-Efficient Specialized Training Regime.}
We further adopt specialized training designs for the above two models.
The structure generator $\epsilon_{struct}$, which derives image structures from text, requires training on an extensive dataset encompassing a wide range of concepts. Traditional methods, like Stable Diffusion, often eliminate low-resolution images, discarding about 48\% of training data and thereby inflating dataset costs. Contrarily, we integrate high-resolution images with upscaled lower-resolution ones. This approach, as proven by our ablated experiments in Sec. \ref{sec:ablation_study}, shows no performance drop, as the predicted $z_{T_{struct}}$ still contains substantial noise. In this way, we achieve higher data efficiency and avoid the problem of semantic degeneration.

Additionally, since the image structure is determined in $z_{T_{struct}}$ and the texture generator $\epsilon_{texture}$ focuses on refining texture, we propose training $\epsilon_{texture}$ at a lower resolution while still sampling at high resolution. This strategy, as demonstrated in our ablated experiments in Sec. \ref{sec:ablation_study}, results in no performance drop and no structural problems (e.g., repetitive presentation \cite{variable_size}). Consequently, we achieved an overall 51\% improvement in training efficiency.
Figure \ref{fig:framework} summarizes the data, training, and inference efficiency of different training strategies. Besides higher data and training efficiency, our strategy also achieves higher inference efficiency with fewer inference steps compared to the Cascaded Training strategy and a smaller per-step model compared to the Resolution Boost Training strategy.

\begin{algorithm}[t]
\begin{spacing}{0.85}
\small
\caption{\textbf{Coop-Diffusion}: Multi-Diffusion Fusing}
\label{alg:coop-diffusion-latent-space}
{\textbf{Sub-Module 1. Bridging Latent Space Gap}} \\
\textbf{Input:} random noise $z_T \sim \mathcal{N}(0, I)$, diffusion model $\epsilon_\theta$, decoder $D$, encoder $E$ in latent space A; random noise $z_T' = z_T$, diffusion model $\epsilon_\theta'$, decoder $D'$, encoder $E'$ in latent space B; guidance strength $d$, sampling method $S$.
\begin{algorithmic}[1]
\For{$t=T, \dots, 1$}
    \State $\epsilon_t' = \epsilon_\theta' (z_t')$, $\hat{z}_{0,t}' = \frac{1}{\sqrt{\bar{\alpha}_t}} (z_t' - \sqrt{1 - \bar{\alpha}_t} \epsilon_t')$
    \State $\hat{x}_{0,t}' = D'(\hat{z}_{0,t}')$, $\tilde{z}_{0,t} = E(\hat{x}_{0,t}')$
    \State $\tilde\epsilon_t = \frac{1}{\sqrt{1 - \bar{\alpha}_t}} (z_t -  \sqrt{\bar{\alpha}_t} {\tilde z}_{0,t})$
    \State $\epsilon_t = \epsilon_\theta (z_t)$,  ${\epsilon_{t,fuse}} = d \cdot \tilde\epsilon_t + (1-d) \cdot \epsilon_t$
    \State $z_{t-1} = S(z_t, t, \epsilon_{t,fuse})$
    \State $z_{t-1}' = S(z_t', t, \epsilon_{t,fuse}')$ \Comment{$\epsilon_{t,fuse}'$ from $\epsilon_{t,fuse}$ similar to the process from $\epsilon_t'$ to $\tilde\epsilon_t$, omitted for brevity}
\EndFor
\State \Return $D(z_0)$
\end{algorithmic}
{\textbf{Sub-Module 2. Bridging Resolution Gap}} \\
\textbf{Input:} diffusion model $\epsilon_\theta$, decoder $D$, encoder $E$ in high-resolution space; random noise $z_T' \sim \mathcal{N}(0, I)$, diffusion model $\epsilon_\theta'$, decoder $D'$, encoder $E'$ in low-resolution space; low-resolution sampling end step $T_{low}$, sampling method $S$.
\begin{algorithmic}[1]
\For{$t=T, \dots, T_{low}+1$}
    \State $\epsilon_t' = \epsilon_\theta' (z_t')$, $z_{t-1}' = S(z_t', t, \epsilon_t')$
\EndFor
\State $\hat{z}_{0,T_{low}}' = \frac{1}{\sqrt{\bar{\alpha}_{T_{low}}}} (z_{T_{low}}' - \sqrt{1 - \bar{\alpha}_{T_{low}}} \epsilon_{T_{low}}')$
\State $\hat{x}_{0,T_{low}}' = D'(\hat{z}_{0,T_{low}}')$, $\hat{x}_{0,T_{low}} = $Upsample$(\hat{x}_{0,T_{low}}')$
\State $\hat{z}_{0,T_{low}} = E(\hat{x}_{0,T_{low}})$
\State $z_{T_{low}} = \sqrt{\bar{\alpha}_{T_{low}}}\hat{z}_{0,T_{low}} + \sqrt{1-\bar{\alpha}_{T_{low}}} \epsilon$, $\epsilon \sim \mathcal{N}\left(0, I\right)$
\For{$t=T_{low}, \dots, 1$}
    \State $\epsilon_t = \epsilon_\theta (z_t)$, $z_{t-1} = S(z_t, t, \epsilon_t)$
\EndFor
\State \Return $D(z_0)$
\end{algorithmic}
\end{spacing}
\end{algorithm}

\begin{figure}[!t]
\vspace{-1mm}
\centering
\includegraphics[width=0.9\linewidth]{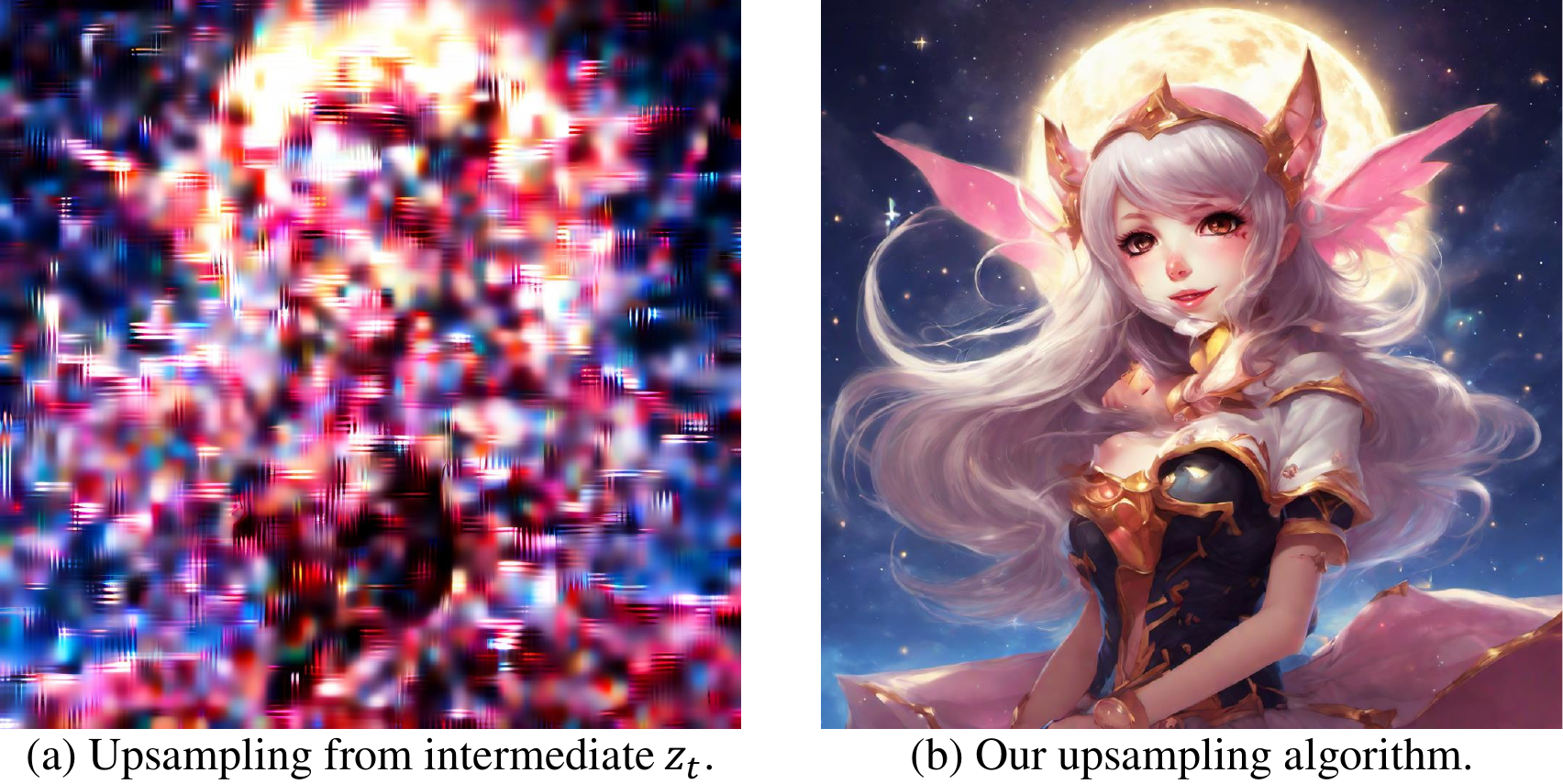}
\vspace{-3mm}
\caption{Results of fusing a low-resolution model and a high-resolution model with different upsampling methods. Upsampling from intermediate $z_t$ results in severe artifacts, while our upsampling algorithm results in high-fidelity image.}
\label{fig:resolution_gap}
\vspace{-3mm}
\end{figure}

\vspace{-1mm}
\subsection{{Coop-Diffusion}: Multi-Diffusion Fusion}
\vspace{-1mm}
As shown in Figure \ref{fig:coop-diffuse}(a), there are numerous pre-trained diffusion models, such as various SD, ControlNet, image variation, etc., each tailored for specific controls and image resolutions. It is promising to fuse these pre-trained models for multi-control or multi-resolution image generation without needing to train a new model. However, the different latent spaces and resolutions of these models impede joint synthesis of images controlled by different models, thereby limiting their practical applications. In response to these challenges, we propose the \textit{Coop-Diffusion} algorithm with two key sub-modules, as shown in Figures \ref{fig:coop-diffuse}(b) and (c), to bridge the latent space gap and the resolution gap, and to unite the denoising process in the same space.

\noindent \textbf{Bridging the Latent Space Gap.}
To bridge the latent space gap between spaces A and B, we propose to unify the model prediction in latent space A by transforming the model prediction $\epsilon_t'$ in latent space B to latent space A using the image space as an intermediate. This is done in the following way: first, we predict the clean data $\hat{z}_{0,t}'$ using Equation (\ref{eq:epsilon_to_x0}) as:
\begin{small}
\begin{equation}
\label{eq:epsilon_to_x0_concrete}
\hat{z}_{0,t}' = \frac{1}{\sqrt{\bar{\alpha}_t}} (z_t' - \sqrt{1 - \bar{\alpha}_t} \epsilon_t'),
\end{equation}
\end{small}
which is then decoded into a pixel-level image $\hat{x}_{0,t}'$ using the latent decoder model $D'$. This image is encoded into latent space A using the image encoder model $E$, as $\tilde{z}_{0,t} = E(\hat{x}_{0,t}')$, and finally transformed into a model prediction by inverting Equation (\ref{eq:epsilon_to_x0}) as:
\begin{small}
\begin{equation}
\tilde\epsilon_t = \frac{1}{\sqrt{1 - \bar{\alpha}_t}} (z_t - \sqrt{\bar{\alpha}_t} \tilde{z}_{0,t}).
\end{equation}
\end{small}
With the united $\tilde\epsilon_t$, we can now perform multi-control fusion between $\tilde\epsilon_t$ and $\epsilon_t$ (the prediction from model $\epsilon_\theta$ with $z_t$ in latent space A, omitted in Figure \ref{fig:coop-diffuse} for brevity) as: ${\epsilon_{t,fuse}} = d \cdot \tilde\epsilon_t + (1-d) \cdot \epsilon_t$, where $d$ and $1-d$ are the guidance strengths of each model with $d \in [0, 1]$, to guide the denoising process jointly with these two models for multi-control image generation. Algorithm \ref{alg:coop-diffusion-latent-space} further illustrates this fusion process.

\noindent \textbf{Bridging Resolution Gap.}
To integrate the denoising processes of a low-resolution model with a high-resolution model, upsampling and/or downsampling is necessary. Traditional bilinear upsampling, often applied to the intermediate result $z_t$ during the denoising process, can undesirably amplify pixel correlation. This amplification deviates from the initial Independent and Identically Distributed (IID) assumption, leading to severe artifacts in the final images, as shown in Figure \ref{fig:resolution_gap}(a). Conversely, downsampling does not present this issue. To address the IID issue in upsampling, we propose a new upsampling algorithm that preserves the IID assumption, thereby bridging the resolution gap between models with different pre-trained resolutions.

Figure \ref{fig:coop-diffuse}(c) visualizes our upsampling algorithm. Specifically, for a low-resolution $z_t'$, we use the image space as an intermediate space to transform $z_t'$ in low-resolution space into high-resolution space as $\tilde z_t$. We first predict the noise $\epsilon_t'$ with the denoising model $\epsilon_\theta'$ and then predict the clean data $\hat{z}_{0,t}'$ as described in Eq. \ref{eq:epsilon_to_x0_concrete}. This is decoded into an image $\hat{x}_{0,t}'$ using decoder $D'$. We then perform upsampling on $\hat{x}_{0,t}'$ to obtain its high-resolution counterpart $\hat{x}_{0,t}$. Finally, $\hat{x}_{0,t}$ is encoded into the latent space with encoder $E$ as $\hat{z}_{0,t}$, and $t$-step noise is added to get the final result $\tilde z_t$ using Eq. \ref{eq:q_sample}.

With the unified $\tilde z_t$, we can now perform multi-resolution fusion. First, we denoise with a low-resolution model to obtain the intermediate $z_t'$ and its high-resolution counterpart $\tilde z_t$. Then, we perform denoising with a high-resolution model starting from $\tilde z_t$, and vice versa. This approach allows us to conduct one-stage super-resolution without undergoing all the low-resolution denoising steps, thereby improving inference efficiency. Algorithm \ref{alg:coop-diffusion-latent-space} further illustrates this fusion process.

%% file: arxiv_sec/4_experiments.tex
\vspace{-1mm}
\section{Experiments}
\vspace{-1mm}
\label{sec:experiments}

\begin{figure*}[!t]
\vspace{-8mm}
\centering
\includegraphics[width=0.9\linewidth]{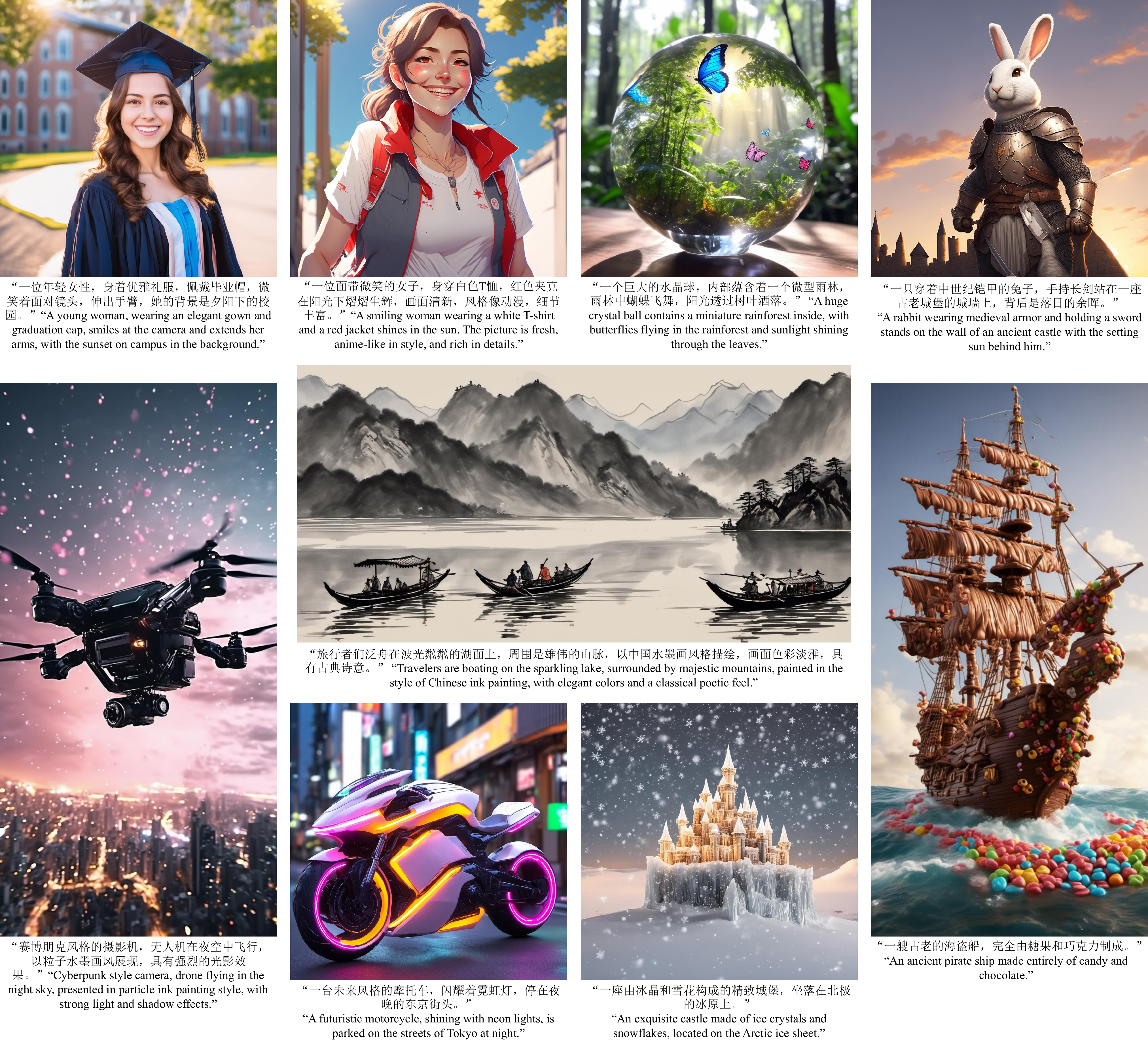}
\vspace{-3mm}
\caption{Images generated with PanGu-Draw, our 5B multi-lingual text-to-image generation model. PanGu-Draw is able to generate multi-resolution high-fidelity images semantically aligned with the input prompts.}
\vspace{-3mm}
\label{fig:generation_results}
\end{figure*}

\begin{table}[!t]
\centering
\caption{Comparisons of PanGu-Draw with recent representative English text-to-image generation models on COCO dataset in terms of FID. Our classifier-free guidance scale is set as 2.}
\vspace{-3mm}
\begin{tabular}{cccc}
\hline
Method       & FID$\downarrow$  & Model Size & Release \\ \hline
DALL-E \cite{dalle} & 27.50 & 12B & N \\
LDM \cite{sd} & 12.63 & 1.5B & \textcolor{red}{Y}\\
GLIDE \cite{nichol2021glide} & 12.24 & 5B & N\\
SDXL \cite{podell2023sdxl} & 11.93 & 2.5B & \textcolor{red}{Y}\\
PixArt-$\alpha$ \cite{chen2023pixartalpha} & 10.65 & 0.6B & \textcolor{red}{Y} \\
DALL-E 2 \cite{ramesh2022dalle2} & 10.39 & 5.5B & N \\
Imagen \cite{imagen} & 7.27 & 3B & N\\
RAPHAEL \cite{xue2023raphael}  & \textbf{6.61} & 3B & N \\ \hline
PanGu-Draw & 7.99 & 5B & \textcolor{red}{Y} \\ \hline
\end{tabular}
\label{tab:fid}
\vspace{-3mm}
\end{table}

\looseness=-1
\noindent \textbf{Implementation Details.}
We adopt the pretrained Variational Autoencoder (VAE) model from SDXL \cite{podell2023sdxl},
and we build our structure and texture generator based on the architecture of its U-Net model with the following modifications.
To achieve bilingual text-to-image generation (Chinese and English), we pre-train a Chinese text encoder \cite{gu2022wukong,yao2021filip} on our Chinese training dataset. We then concatenate the text embeddings from this Chinese text encoder with those from a pretrained English text encoder, serving as the final text embeddings for the denoising models.
For multi-resolution image generation, we select a range of image resolutions around 1024x1024 and further condition the denoising model on the sinusoidal positional embeddings corresponding to the index of image resolutions. The $T_{struct}$ parameter is set to 500, as suggested by our ablation study.

Our models are trained on a cluster consisting of 256 Ascend 910B cards. During training, we applied several techniques to reduce redundant memory usage. These include replacing traditional attention with Flash Attention \cite{dao2022flashattention}, employing mixed-precision training \cite{micikevicius2018mixed}, and using gradient checkpointing \cite{chen2016training}, also known as the recompute technique. These methods enable the model to fit within the memory of a single Neural Processing Unit (NPU), allowing parallelism to be applied only in the data scope and avoiding model sharding among NPUs, as well as reducing inter-machine communication overhead.

\noindent \textbf{Dataset Construction.}
To encompass the abundant concepts in the world, we collect images in various styles from multiple sources, including Noah-Wukong \cite{gu2022wukong}, LAION \cite{rombach2022high}, and others, such as photography, cartoons, portraits, and gaming assets.
The collected images are filtered based on CLIP score, aesthetic score, watermark presence, resolution, and aspect ratio. To improve the semantic alignment of PanGu-Draw, we discard parts of the noisy captions that are meaningless or mismatched to the image, sourced from the Internet. Instead, we recaption the collected images by first employing an open-vocabulary detector \cite{yao2023detclipv2} to locate the primary subjects within the images. These subjects are then processed by LLaVA \cite{liu2023visual}, a high-performance vision-language model, along with prompting templates, to yield detailed image descriptions. These English annotations are subsequently translated into Chinese.

\noindent \textbf{Evaluation Metrics.} We evaluate PanGu-Draw's text-to-image generation on COCO  \cite{coco} with 30k  images for English, and COCO-CN \cite{li2019cococn} with 10k images for Chinese. The Frechet Inception Distance (FID \cite{fid}) is utilized to evaluate image quality and diversity. For Chinese, additional metrics include the Inception Score (IS \cite{salimans2016IS}) and CN-CLIP-score\cite{chinese-clip}, assessing image quality and text-image alignment. Complementing these metrics, a user study is conducted to evaluate image-text alignment, fidelity, and aesthetics using ImageEval-prompt\footnote{https://github.com/FlagOpen/FlagEval/tree/master/imageEval} across 339 prompts.

\vspace{-1mm}
\subsection{Text-to-Image Generation}
\vspace{-1mm}
\noindent \textbf{Evaluation on COCO.}
As shown in Table \ref{tab:fid}, PanGu-Draw achieves a FID of 7.99, which is superior to compared methods such as DALL-E 2 and SDXL. It also achieves competitive FID with SOTA methods, indicating the effectiveness of our time-decoupling training strategy and its outstanding data and training efficiencies. Our 5B PanGu model is the best-released model in terms of FID.

\noindent \textbf{Evaluation on COCO-CN.}
As shown in Table \ref{tab:cococn}, PanGu-Draw outperforms other released Chinese text-to-image models, including Taiyi-CN, Taiyi-Bilingual, and AltDiffusion, across all three metrics. This performance highlights PanGu-Draw's exceptional Chinese text-to-image generation capabilities and the effectiveness of our bilingual text encoder architecture.

\begin{table}[!t]
\vspace{-4mm}
\centering
\caption{Comparisons of PanGu-Draw with Chinese text-to-image generation models on COCO-CN dataset in terms of FID, IS and CN-CLIP-score. The classifier-free guidance scales are set as 9 following AltDiffusion \cite{ye2023altdiffusion}.}
\vspace{-3mm}
\begin{tabular}{cccc}
\hline
Model           & FID$\downarrow$   & IS$\uparrow$    & CN-CLIP-score$\uparrow$ \\ \hline
AltDiffusion\cite{ye2023altdiffusion}    & 25.31 & 29.16 & 35.12         \\
Taiyi-Bilingual\cite{fengshenbang} & 24.61 & 34.29 & 32.26         \\
Taiyi-CN\cite{fengshenbang}        & 23.99 & 34.29 & 34.22         \\ \hline
PanGu-Draw      & \textbf{21.81} & \textbf{37.00}    & \textbf{36.62}         \\ \hline
\end{tabular}
\label{tab:cococn}
\vspace{-2mm}
\end{table}

\begin{table}[!t]
\vspace{-1mm}
\centering
\caption{Results of a User study on ImageVal-prompt in terms of image-text alignment, image fidelity, and aesthetics.}
\vspace{-3mm}
\resizebox{\linewidth}{!}{
\begin{tabular}{ccccc}
\hline
Method         & Align$\uparrow$ & Fidelity$\uparrow$ & Aesthetics$\uparrow$ & Ave$\uparrow$ \\ \hline
DALL-E 3\cite{dalle3} & \textbf{4.72} & \textbf{4.59} & \textbf{4.76}  & \textbf{4.69}\\
MJ 5.2 &        4.63              &    4.54    & 4.75 & 4.64       \\
SDXL\cite{podell2023sdxl}           &    4.41                  &    4.37  &   4.59 & 4.46      \\ 
SD\cite{sd} & 4.17 & 3.99 & 4.20 & 4.12 \\ \hline
PanGu-Draw     &         4.5             &    4.52   & 4.72 & 4.58       \\ \hline
\end{tabular}
}
\label{tab:human_evaluation}
\vspace{-2mm}
\end{table}

\noindent \textbf{User Study.}
We conducted a user study to compare PanGu-Draw with top-performing methods, including SDXL \cite{podell2023sdxl}, Midjourney 5.2, and DALL-E 3 \cite{dalle3}. As shown in Table \ref{tab:human_evaluation}, PanGu-Draw achieves better results than SD and SDXL across all three metrics. It also attains approximately 99\%/98\% of the performance of Midjourney 5.2 and DALL-E 3, respectively, indicating PanGu-Draw's excellent text-to-image capabilities. Figure \ref{fig:generation_results} shows a collection of high-fidelity multi-resolution images generated by PanGu-Draw. As we can see, the generated images of PanGu-Draw are of high aesthetics and semantically aligned with the input prompts.

\begin{figure}[!t]
\vspace{-4mm}
\centering
\includegraphics[width=0.9\linewidth]{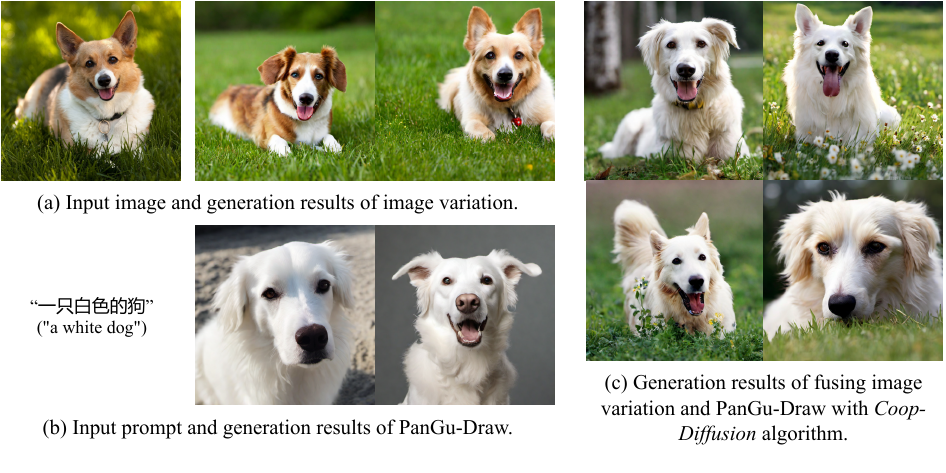}
\vspace{-6mm}
\caption{Generation results of the fusing of an image variation model and PanGu-Draw and with the proposed \textit{Coop-Diffusion} algorithm. } 
\label{fig:multicontrol_text_image}
\vspace{-2mm}
\end{figure}

\begin{figure}[!t]
\centering
\includegraphics[width=0.9\linewidth]{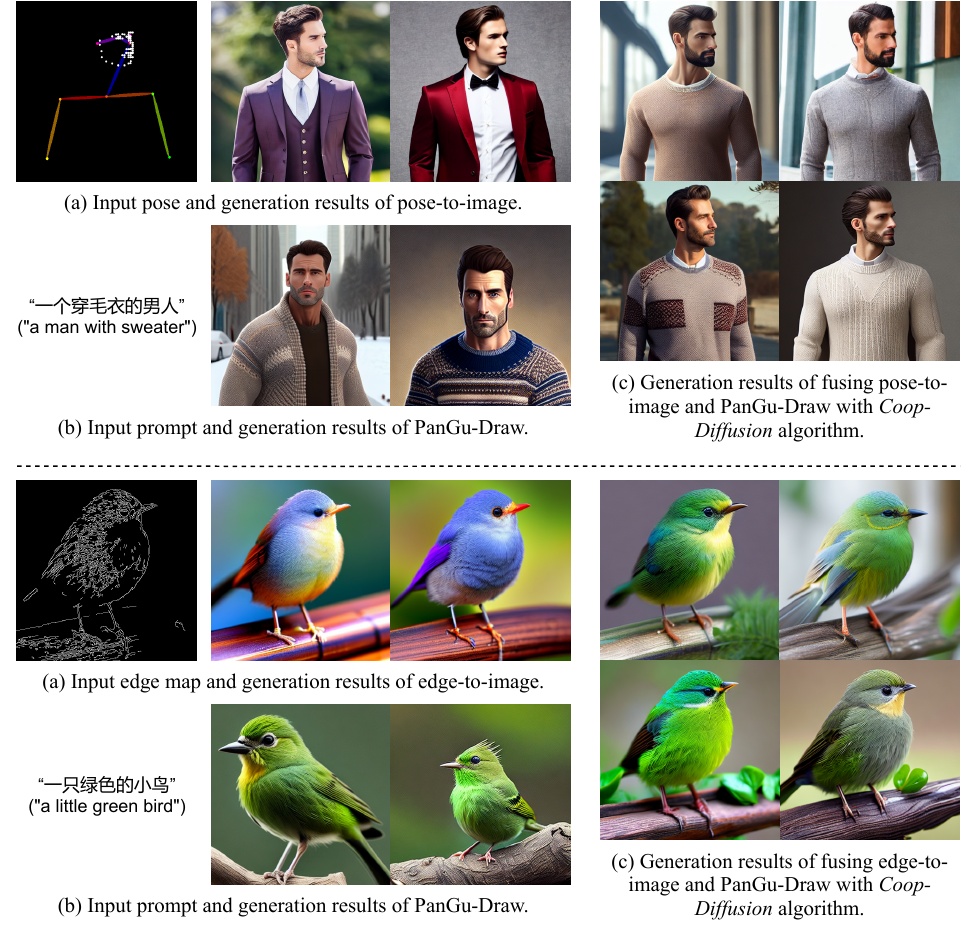}
\vspace{-6mm}
\caption{Generation results guided by fusing signals of text and pose/edge map by our \textit{Coop-Diffusion}.}
\vspace{-3mm}
\label{fig:multicontrol_text_pose}
\end{figure}

\vspace{-1mm}
\subsection{Multi-Diffusion Fusing Results}
\vspace{-1mm}

\noindent \textbf{Multi-Control Image Generation.}
To demonstrate the effectiveness of the proposed reusable multi-diffusion fusing algorithm, \textit{Coop-Diffusion}, we first present multiple results of multi-control image generation. Figure \ref{fig:multicontrol_text_image} displays results from fusing an image variation model \footnote{https://huggingface.co/lambdalabs/sd-image-variations-diffusers} with PanGu-Draw. The fusing results maintain a style similar to that of the reference image, matching the texture described by the input prompt. Figure \ref{fig:multicontrol_text_pose} shows results from fusing PanGu-Draw with a pose/edge-to-image ControlNet model, which operates in guess mode without input prompts. Here, the fusing results combine the structure of the pose/edge image with the texture described by the input prompt.

\begin{figure}[!t]
\vspace{-6mm}
\centering
\includegraphics[width=0.8\linewidth]{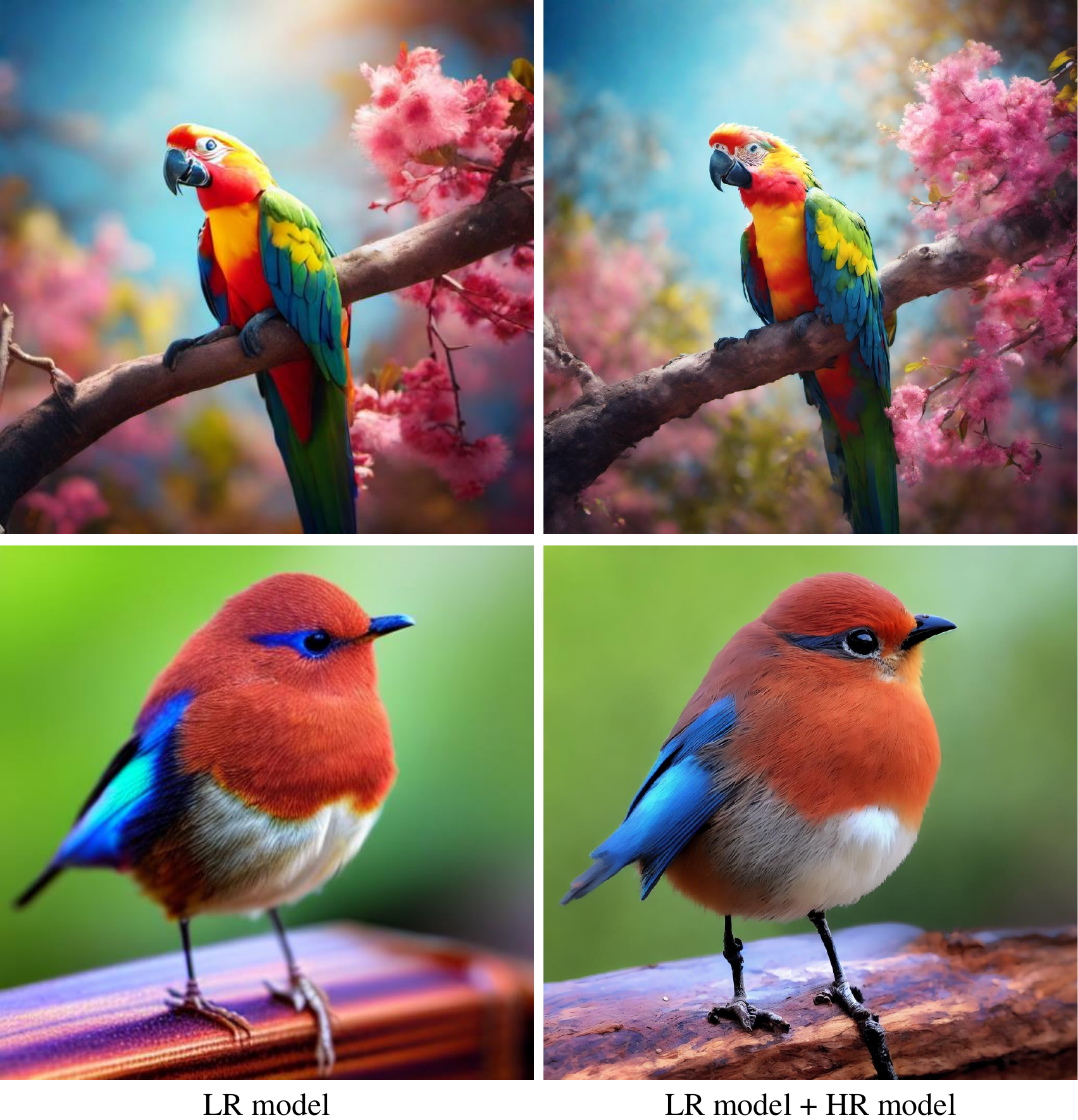}
\vspace{-3mm}
\caption{Images generated with a low-resolution (LR) model (first row: text-to-image model; second row: Edge-to-image ControlNet) and the fusion of the LR model and HR PanGu-Draw with our \textit{Coop-Diffusion}. This allows for single-stage super-resolution for better details and higher inference efficiency.}
\vspace{-3mm}
\label{fig:coop-diffusion-2}
\end{figure}

\noindent \textbf{Multi-Resolution Image Generation.}
We also present multi-resolution image generation results of fusing PanGu-Draw with low-resolution text-to-image and edge-to-image ControlNet model by first denoising with the low-resolution model to get the intermediate $z_t$ and the high-resolution counterpart $\tilde z_t$, and then perform denoising in high resolution with PanGu-Draw. Figure \ref{fig:coop-diffusion-2} shows the results from the low-resolution model and our fusing algorithm \textit{Coop-Diffusion}. As we can see, PanGu-Draw adds much details to the low-resolution predictions leading to high-fidelity high-resolution results. Besides, compared with the common practice of super-resolution with diffusion model, which carries out all the low-resolution denoising steps, our method achieve higher inference efficiency.

\vspace{-1mm}
\subsection{Ablation Study}
\vspace{-1mm}
\label{sec:ablation_study}
In this section, we perform ablation studies to analyze our time-decoupling training strategy.
The baseline model has $1B$ parameters while the structure and texture generators both have $0.5B$ parameters. During the training process, the latter two models only train half the steps of the baseline model with $T_{struct}$ set as 500. Both settings of the models are trained from scratch on a subset 
of the LAION dataset containing images with all sizes. After training, FID, IS and CLIP-score on COCO are reported for comparison.

\noindent \textbf{Time-Decoupling Training Strategy.}
We compare the final performance of models trained with the Resolution Boost strategy and our time-decoupling strategy in Table \ref{tab:model_splitting}. We found that models trained with our strategy achieves much better performance in all three criteria, indicating the effectiveness of our strategy.

\begin{table}[!t]
\vspace{-6mm}
\centering
\caption{Comparison of models across Resolution Boost (1B parameters) and Time-Decoupling training strategies (0.5B parameters for structure and texture generators)}
\vspace{-3mm}
\begin{tabular}{cccc}
\hline
Model              & FID$\downarrow$ & IS$\uparrow$ & CLIP-score$\uparrow$ \\ \hline
Resolution Boost &  106.12   & 10.46   &  22.9          \\
Time-Decoupling &  \textbf{87.66}   &  \textbf{11.07}  &  \textbf{23.4 }         \\ \hline
\end{tabular}
\label{tab:model_splitting}
\end{table}

\begin{table}[!t]
\vspace{-3mm}
\centering
\caption{Performance of structure and texture models training with images of different resolutions.}
\vspace{-3mm}
\resizebox{\columnwidth}{!}{
\begin{tabular}{ccccc}
\hline
 Structure & Texture              & \multirow{2}{*}{FID$\downarrow$} & \multirow{2}{*}{IS$\uparrow$} & \multirow{2}{*}{CLIP-score$\uparrow$} \\ 
 Data & Resolution & & & \\
 \hline
 All data & 256  &  \textbf{87.66}   &  \textbf{11.07}  &  \textbf{23.4 }         \\ \hline
 Only high resolution & 256  &  89.52   &  10.96  &  23.2         \\ 
 All data & 512 & 90.98 & 10.59 & 23.3 \\
 \bottomrule
\end{tabular}
}
\label{tab:training_designs}
\vspace{-1mm}
\end{table}

\noindent \textbf{Training Designs.}
The structure and texture generators ($\epsilon_{\text{struct}}$ and $\epsilon_{\text{texture}}$) are designed to train on different resolutions to improve data and training efficiency. However, this approach may negatively influence the final performance. In Table \ref{tab:training_designs}, we compare such a design with a traditional training process, where $\epsilon_{\text{struct}}$ discards low-resolution images, or $\epsilon_{\text{texture}}$ trains with high resolution. Results on COCO show that $\epsilon_{\text{struct}}$ benefits from these extra up-scaled data, and $\epsilon_{\text{texture}}$ learns enough texture patterns at a smaller resolution. 

\begin{table}[!t]
\vspace{-1mm}
\centering
\caption{Comparisons of PanGu-Draw inference performance with different time step splitting point $T_{struct}$ settings. }
\vspace{-3mm}
\begin{tabular}{cccc}
\hline
$T_{struct}$ & FID$\downarrow$                  & IS$\uparrow$                   & CLIP-score$\uparrow$           \\ \hline
200 & 105.08 & 10.59 & 22.98 \\
300      & 98.08                 &          10.72                            &          23.12            \\
500      & \textbf{87.66}               & \textbf{11.07} & \textbf{23.40 } \\
700                       &    89.48                  &         11.02             &   23.32                   \\ \hline
\end{tabular}
\label{tab:time_step_splitting_point}
\vspace{-5mm}
\end{table}

\noindent \textbf{Timestep Splitting Point.}
The timestep splitting point $T_{\text{struct}}$ between the structure and texture generators also influences the final performance. To this end, we set $T_{\text{struct}}$ to 200, 300, 500, and 700, while keeping the other settings of the structure and texture generators unchanged. As shown in Table \ref{tab:time_step_splitting_point}, as $T_{\text{struct}}$ increases from 200 to 700, the performance initially increases and then decreases continuously. $T_{\text{struct}} = 500$ is the optimal value, and we adopt it as the default setting in all other experiments.

%% file: arxiv_sec/5_conclusion.tex
\vspace{-1mm}
\section{Conclusion}
\vspace{-1mm}
\label{sec:experiments}
In this paper, we present ``PanGu-Draw'', a new latent diffusion model for efficient text-to-image generation that effectively integrates multiple control signals. Our approach includes a Time-Decoupling Training Strategy to separate the text-to-image process into structure and texture generation, enhancing data use and computational efficiency. Additionally, ``Coop-Diffusion'' is introduced, an algorithm allowing cooperative use of different pre-trained diffusion models in a unified denoising process for multi-control image synthesis at various resolutions without extra data or retraining. 
PanGu-Draw outperforms models like DALL-E 2 and SDXL in English T2I, achieves superior FID, IS, and CN-CLIP-scores in Chinese T2I, and receives favorable user feedback. 
This positions PanGu-Draw as a versatile and efficient state-of-the-art method, which is available on the Ascend platform.

%% file: arxiv_sec/X_suppl.tex
\clearpage
\setcounter{page}{1}
\maketitlesupplementary

\section{More Details about PanGu-Draw}

\begin{figure*}[t!]
\centering
\includegraphics[width=1\linewidth]{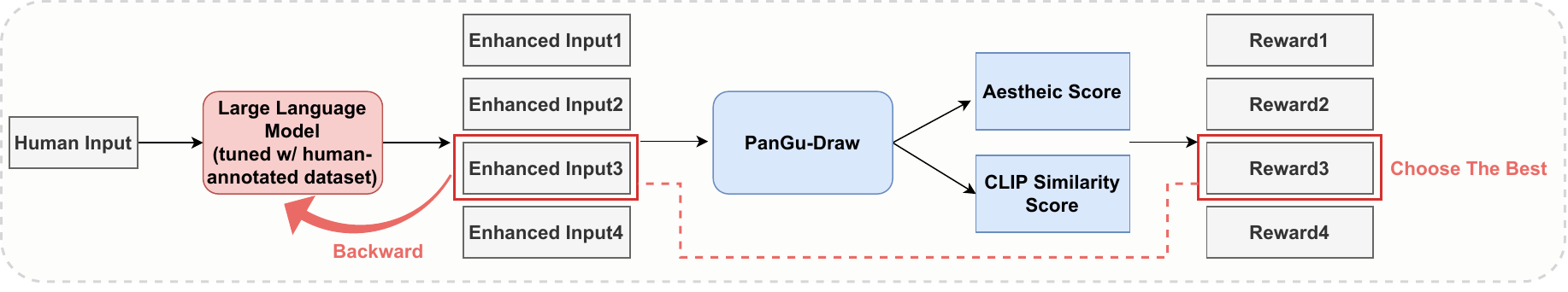}
\caption{Prompt enhancemnent pipeline with Large Language Model~(LLM), specifically tailored for PanGu-Draw. Initially, we fine-tune the LLM using a human-annotated dataset, transforming a succinct prompt into a more enriched version. Subsequently, to optimize for PanGu-Draw, we employ the Reward rAnked FineTuning (RAFT) method, as introduced in \cite{dong2023raft}, which selects the prompt pairs yielding the highest reward for further fine-tuning.}
\label{fig:prompt_enhancement_}
\end{figure*}

\noindent \textbf{Prompt Enhancement LLM with RLAIF Algorithm.}
To further enhance our generation quality, we harness the advanced comprehension abilities of large language models (LLM)\cite{zeng2022glm,du2022glm} to align users' succinct inputs with the detailed inputs required by the model. Specifically, shown in Figure \ref{fig:prompt_enhancement_}, we first construct a human-annotated dataset that enriches succinct prompts with background and style descriptions and then fine-tune the LLM to adapt a succinct prompt to an enriched one using this data. To better adapt to the inputs required by PanGu-Draw, we perform further refinement based on the Reward rAnked FineTuning (RAFT)\cite{dong2023raft} method. Subsequently, we use the fine-tuned LLM to expand on multiple texts, which are then input into PanGu-Draw for image generation. The best expansions are selected jointly by an aesthetic scoring model\footnote{https://github.com/christophschuhmann/improved-aesthetic-predictor} and a CLIP~\cite{radford2021learning} semantic similarity calculation model, allowing for further fine-tuning of the LLM. 

\begin{figure}[!t]
\centering
\includegraphics[width=1\linewidth]{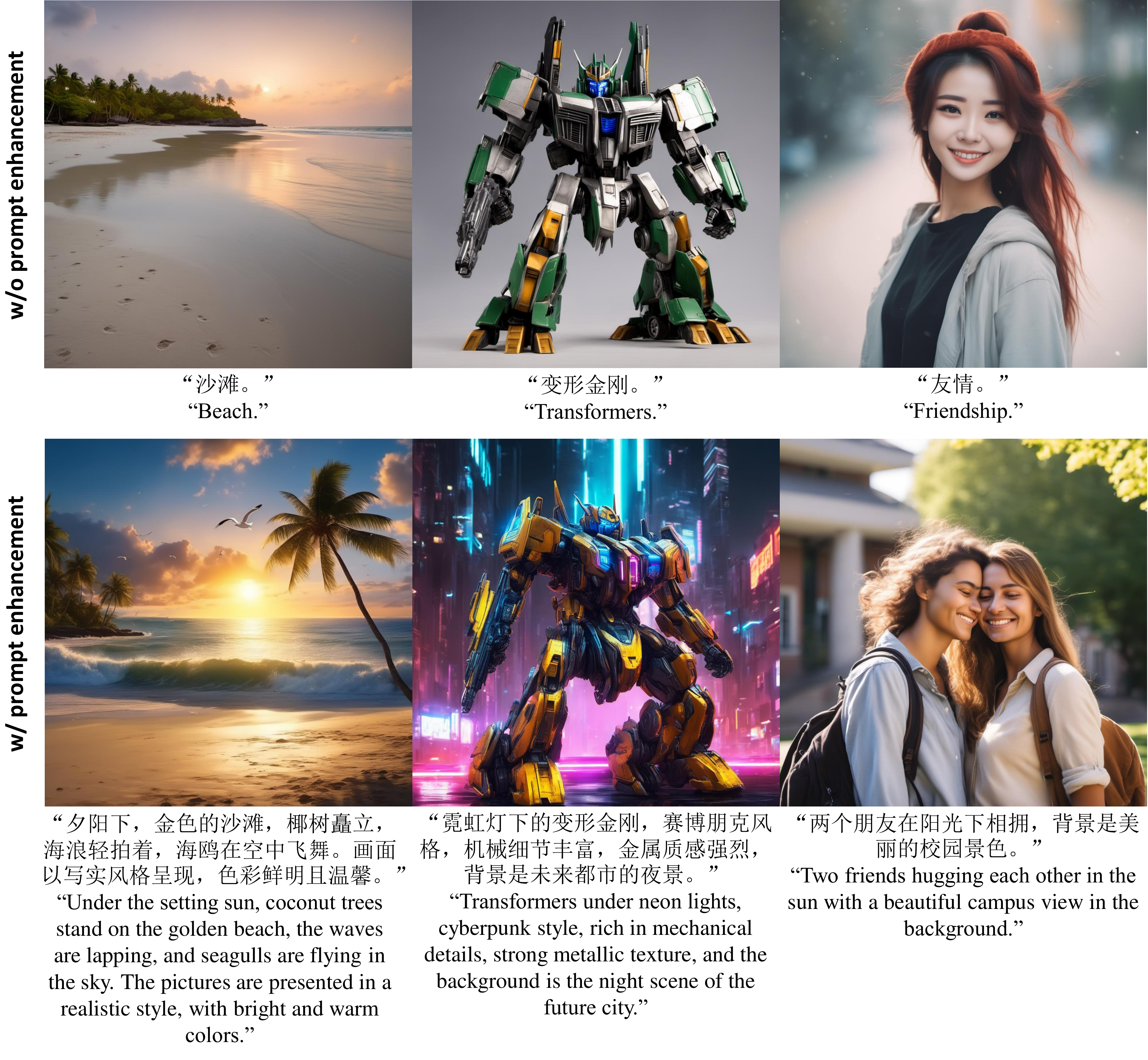}
\caption{Text-to-image generation results without and with prompt enhancement. Enriched text improve image generation by better image aesthetic perception (left), more detailed background (middle) and better interpretation of abstract concepts (right).}
\label{fig:prompt_tuning}
\end{figure}

Figure \ref{fig:prompt_tuning} shows the generation results of PanGu-Draw without and with prompt enhancement. 
As we can see, prompt enhancement serves to add more details and illustration to the original brief prompts, leading to better image aesthetics and semantic alignment. 

\begin{figure}[!t]
\centering
\includegraphics[width=1\linewidth]{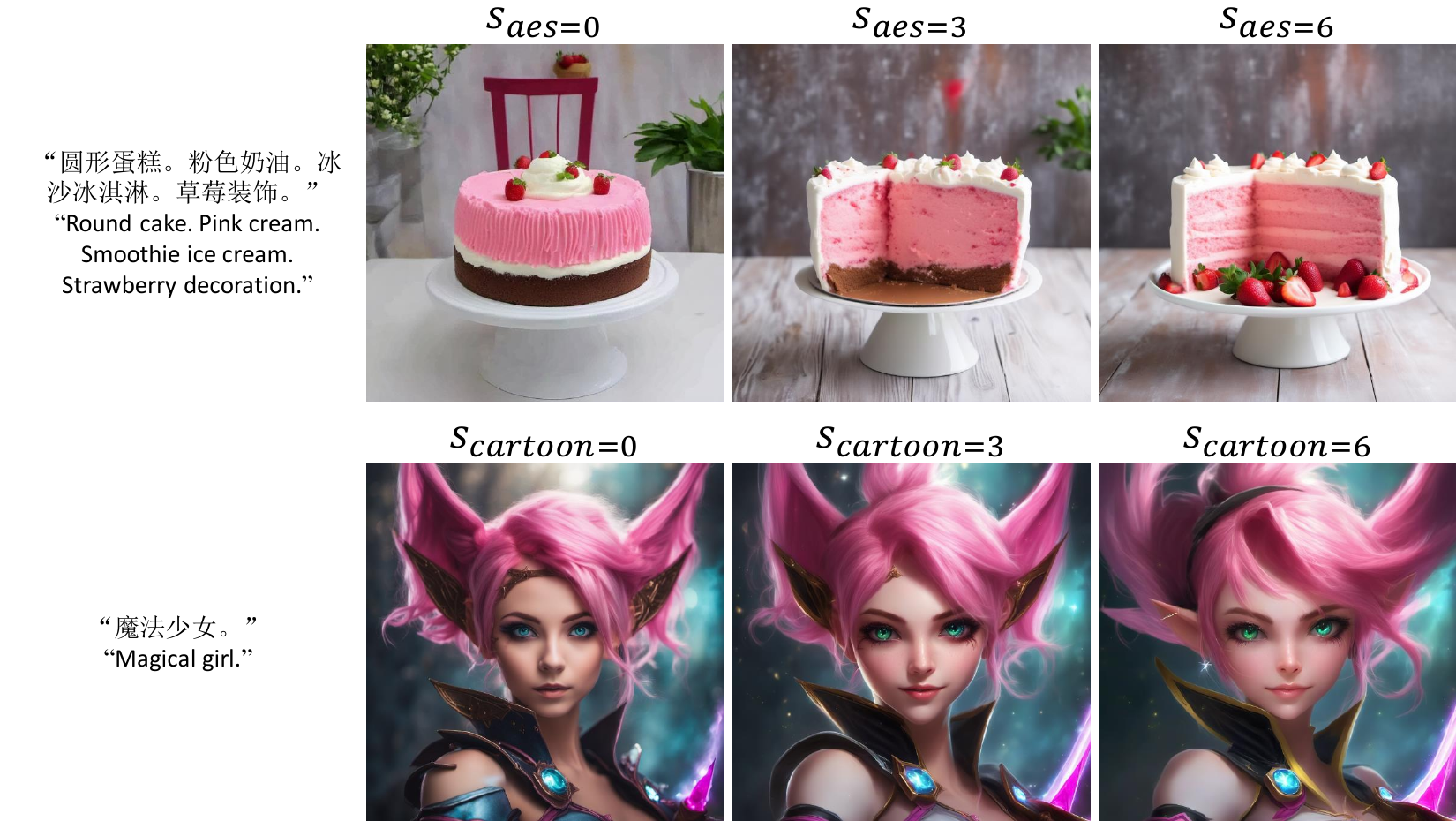}
\caption{Controllable stylized text-to-image generation results of PanGu-Draw. PanGu-Draw can control the generated images towards the desired style with the style guidance scale. $s_{aes}$ for human-aesthetic-prefer style and $s_{cartoon}$ for cartoon style.}
\label{fig:stylized_text_to_image}
\end{figure}

\noindent \textbf{Controllable Stylized Text-to-Image Generation.}
While techniques like LoRA \cite{hu2021lora} allow one to adapt a text-to-image model to a specific style (e.g., cartoon style, human-aesthetic-preferred style), they do not allow one to adjust the degree of the desired style. To this end,
inspired by the classifier-free guidance mechanism, we propose to perform controllable stylized text-to-image generation by first
construct a dataset consisting of human-aesthetic-prefer, cartoon and other samples with a pretrained human aesthetic scoring model and a  cartoon image classification models, and 
then train the text-to-image generation model with these three kinds of samples. For human-aesthetic-prefer and cartoon samples, we prepend a special prefix to the original prompt, denoted as $c_{aes}$ and $c_{cartoon}$ respectively. During sampling, we extrapolated the prediction in the direction of $\epsilon_{\theta}(z_t, t, c_{style})$ and away from $\epsilon_{\theta}(z_t, t, c)$ as follows:
\begin{align*}
    \hat{\epsilon}_{\theta}(z_t, t, c) &= \epsilon_{\theta}(z_t, t, \emptyset) + s \cdot ({\epsilon}_{\theta}(z_t, t, c) - \epsilon_{\theta}(z_t, t, \emptyset)) \\
    &+ s_{style} \cdot ({\epsilon}_{\theta}(z_t, t, c_{style}) - \epsilon_{\theta}(z_t, t, c)),
\end{align*}
where $s$ is the classifier-free guidance scale, $c_{style} \in \{c_{aes}, c_{cartoon}\}$ and $s_{style}$ is the style guidance scale.

Figure \ref{fig:stylized_text_to_image} shows the controllable stylized text-to-image generation results of PanGu-Draw, including human-aesthetic-prefer and cartoon style image generation. As we can see, with the corresponding style guidance scale, PanGu-Draw can control the generated images towards the desired style.

\section{Image Resolutions for Multi-Resolution Training}
Table \ref{tab:resolutions} shows the list of resolutions used for multi-resolution training of our structure generation model and texture generation model.

\begin{table}[!t]
\centering
\caption{The image resolutions used for multi-resolution training of structure generation model and texture generation model.}
\begin{tabular}{cc|cc}
\toprule
\multicolumn{2}{c|}{Structure Generation Model} & \multicolumn{2}{c}{Texture Generation Model} \\ \hline
Height            & Width            & Height           & Width           \\ \hline
512               & 2048             & 256              & 1024            \\
512               & 1920             & 256              & 960             \\
704               & 1408             & 384              & 768             \\
768               & 1344             & 416              & 736             \\
864               & 1152             & 480              & 640             \\
1024              & 1024             & 512              & 512             \\
1152              & 864              & 640              & 480             \\
1344              & 768              & 736              & 416             \\
1408              & 704              & 768              & 384             \\
1920              & 512              & 960              & 256             \\
2048              & 512              & 1024             & 256             \\ \bottomrule
\end{tabular}
\label{tab:resolutions}
\end{table}

\section{More Generation Results of PanGu-Draw}

\subsection{Text-to-Image Generation}

\begin{figure*}[!t]
\centering
\includegraphics[width=1.\linewidth]{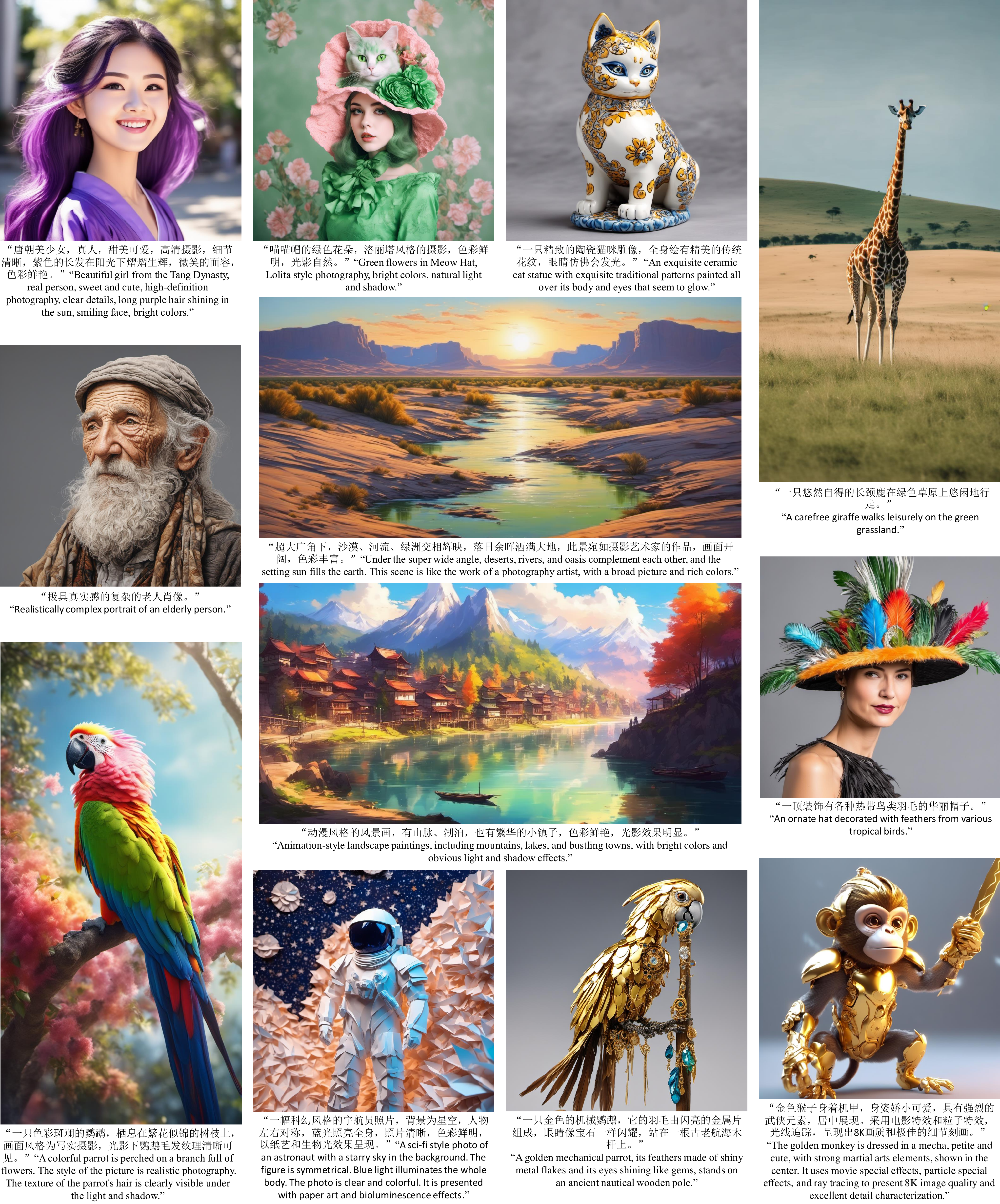}
\caption{Images generated with PanGu-Draw, our 5B multi-lingual text-to-image generation model. PanGu-Draw is able to generate multi-resolution high-fidelity images semantically aligned with the input prompts.}
\label{fig:more_text_to_image}
\end{figure*}

Figure \ref{fig:more_text_to_image} shows more generated images of PanGu-Draw. As we can see, the generated images are of high visual quality and are well aligned with the input prompts.

\begin{figure}[!t]
\centering
\includegraphics[width=1\linewidth]{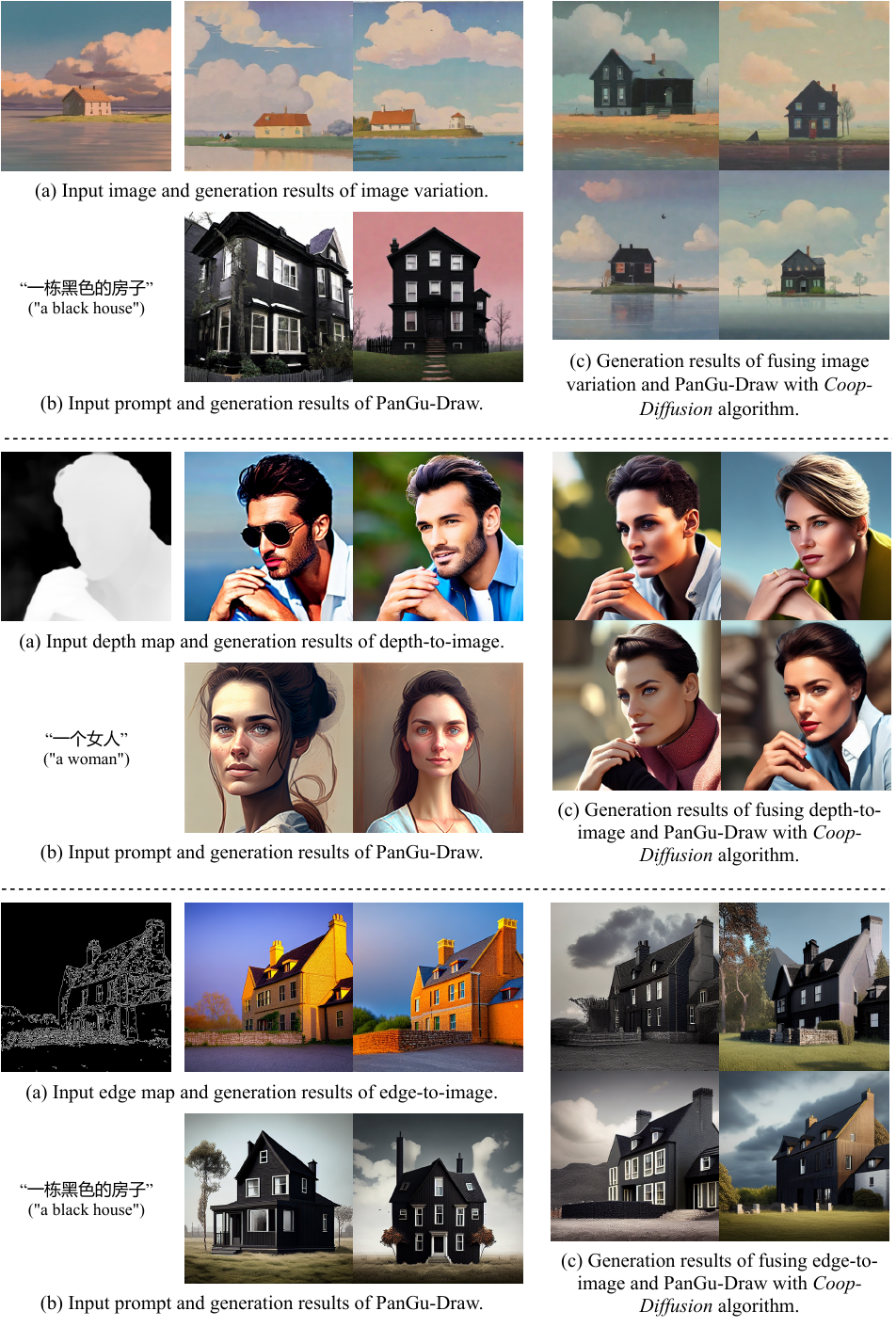}
\caption{Generation results of the fusing of an image variation/depth-to-image/edge-to-image model and PanGu-Draw with the proposed \textit{Coop-Diffusion} algorithm.}
\label{fig:multi-control-more}
\end{figure}

\begin{figure}[!t]
\centering
\includegraphics[width=1\linewidth]{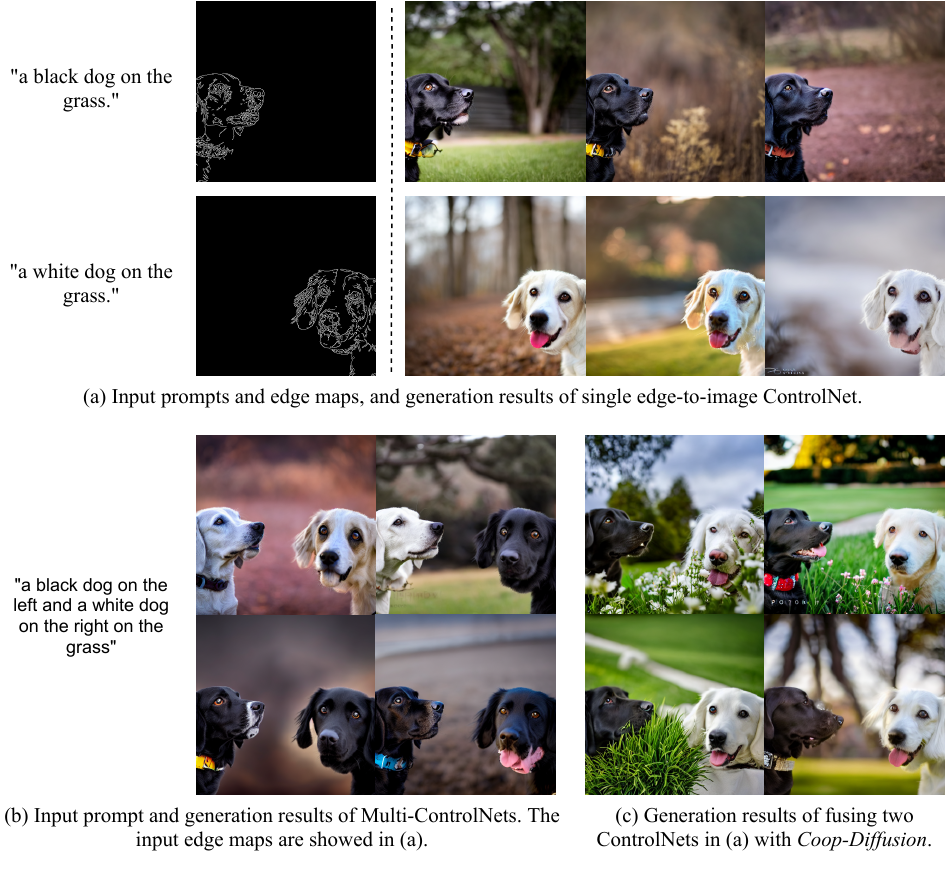}
\caption{Generation results of the fusing of an image variation/depth-to-image/edge-to-image model and PanGu-Draw with the proposed \textit{Coop-Diffusion} algorithm.}
\label{fig:multi-control-2-controlnet}
\end{figure}

\begin{figure}[!t]
\vspace{-3mm}
\centering
\includegraphics[width=1\linewidth]{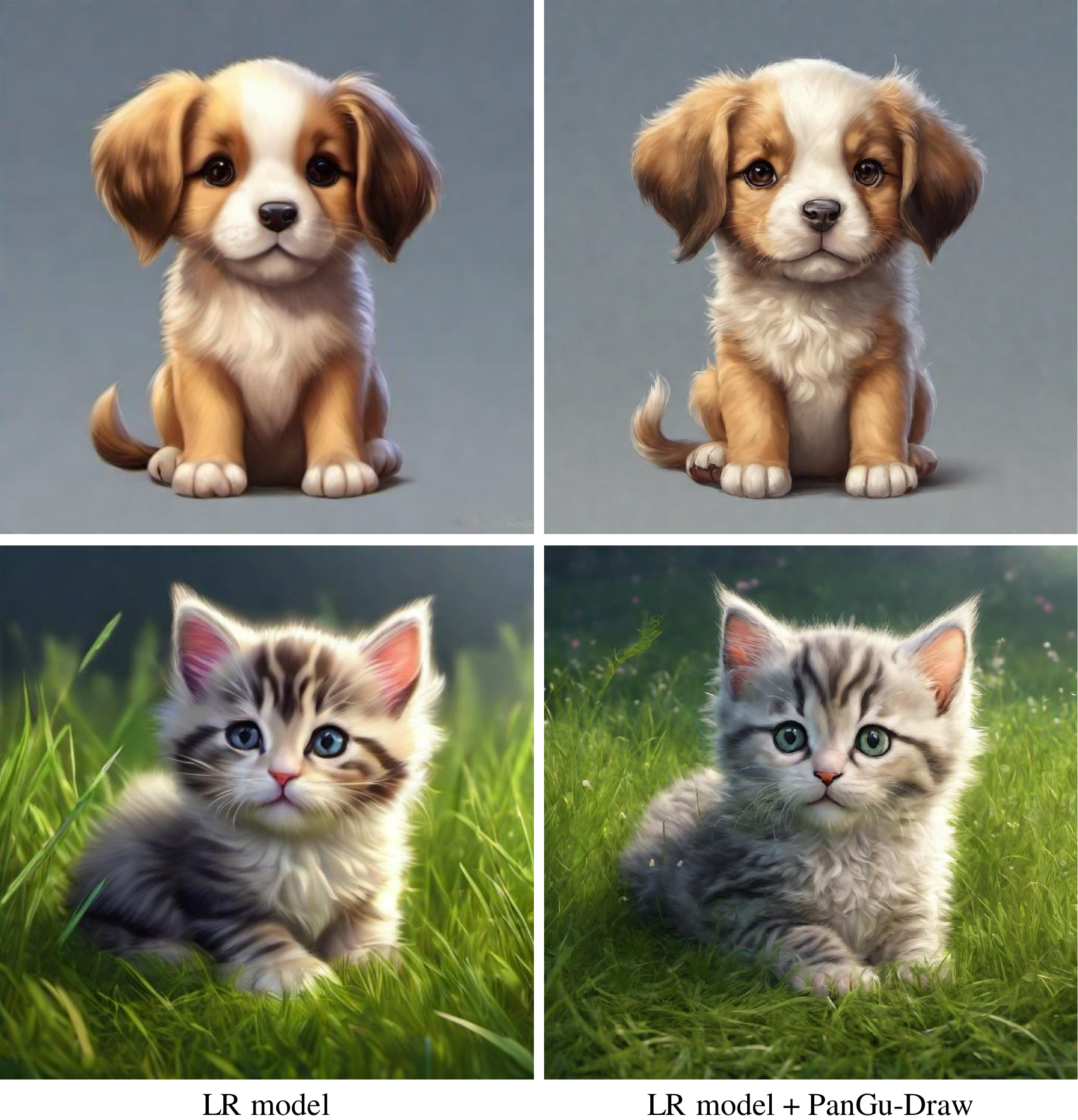}
\vspace{-3mm}
\caption{Images generated with a low-resolution (LR) model and the fusion of the LR model and HR PanGu-Draw with our \textit{Coop-Diffusion}. This allows for single-stage super-resolution for better details and higher inference efficiency.}
\vspace{-3mm}
\label{fig:multi-resolution-2}
\end{figure}

\subsection{Multi-Diffusion Fusing Results}
\noindent \textbf{Multi-Control Image Generation.}
Figure \ref{fig:multi-control-more} shows results of multi-control image generation by fusing PanGu-Draw with different models, including image variation, depth-to-image, edge-to-image generation models.

Figure \ref{fig:multi-control-2-controlnet} shows results of fusing two ControlNet models with our algorithm and with the algorithm proposed by ControlNet \cite{controlnet}, which fuses the features of different ControlNets before injecting into the U-Net model. As we can see, our algorithm is able to specify the prompts of different ControlNets such that enabling a finer-grain control.

\noindent \textbf{Multi-Resolution Image Generation.}
Figure \ref{fig:multi-resolution-2} shows the results from the low-resolution model and our fusing algorithm \textit{Coop-Diffusion} by fusing the low-resolution model and our high-resolution PanGu-Draw model. As we can see, PanGu-Draw adds much details to the low-resolution predictions leading to high-fidelity high-resolution results.

\section{Visual Comparison against Baselines}
\begin{figure*}[!t]
\centering
\includegraphics[width=0.9\linewidth]{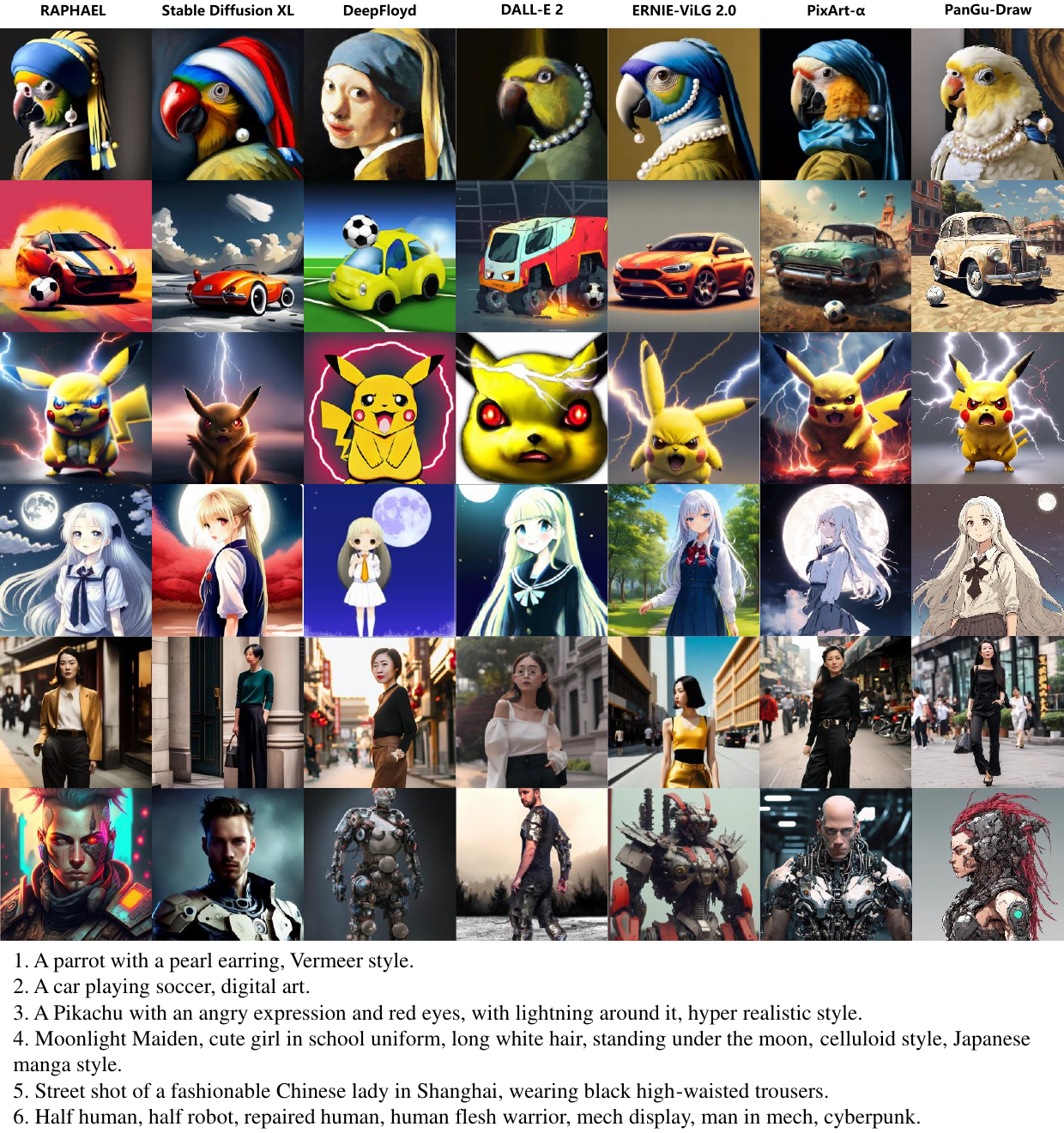}
\caption{Visual comparison of PanGu-Draw against baseline methods, including RAPHAEL \cite{xue2023raphael}, SDXL \cite{podell2023sdxl}, DeepFloyd \cite{shonenkov2023deepfloyd}, DALL-E 2 \cite{ramesh2022dalle2}, ERNIE-ViLG 2.0 \cite{feng2023ernie}, and PixArt-$\alpha$ \cite{chen2023pixartalpha}. The input prompts are also used in RAPHAEL and are provided at the bottom of the figure. The results of PanGu-Draw are better than or on par with these top-performing baseline models.}
\label{fig:visual_compare_1}
\end{figure*}

\begin{figure*}[!t]
\centering
\includegraphics[width=0.9\linewidth]{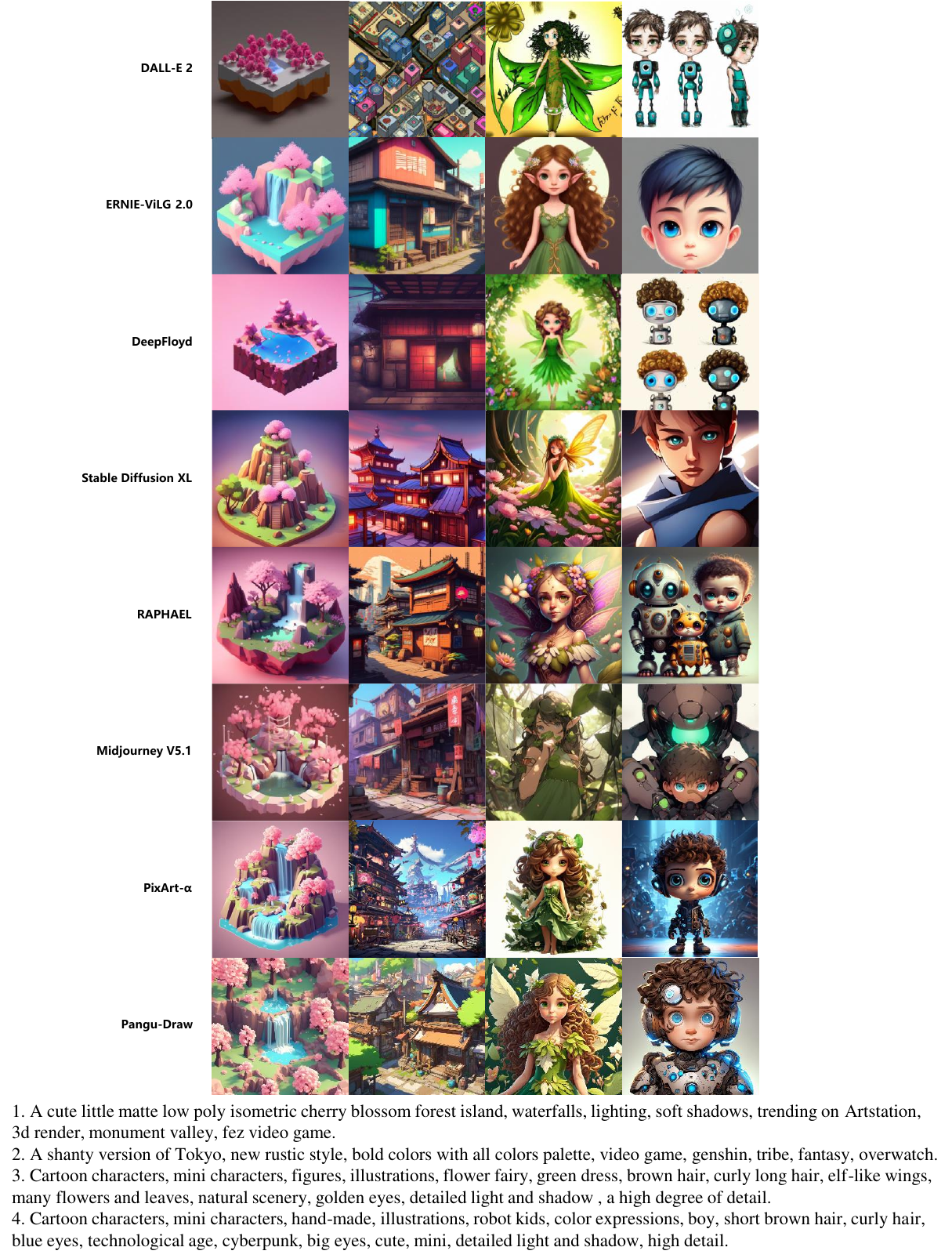}
\caption{Visual comparison of PanGu-Draw against baseline methods, including DALL-E 2 \cite{ramesh2022dalle2}, ERNIE-ViLG 2.0 \cite{feng2023ernie}, DeepFloyd \cite{shonenkov2023deepfloyd}, SDXL \cite{podell2023sdxl}, RAPHAEL \cite{xue2023raphael},Midjourney V5.1 and PixArt-$\alpha$ \cite{chen2023pixartalpha}. The input prompts are also used in RAPHAEL and are provided at the bottom of the figure. The results of PanGu-Draw are better than or on par with these top-performing baseline models.}
\label{fig:visual_compare_2}
\end{figure*}

Figure \ref{fig:visual_compare_1} and \ref{fig:visual_compare_2} shows qualitative comparisons of PanGu-Draw against baseine methods, including RAPHAEL \cite{xue2023raphael}, SDXL \cite{podell2023sdxl}, DeepFloyd \cite{shonenkov2023deepfloyd}, DALL-E 2 \cite{ramesh2022dalle2}, ERNIE-ViLG 2.0 \cite{feng2023ernie}, PixArt-$\alpha$ \cite{chen2023pixartalpha} and . The input prompts are also used in RAPHAEL and are provided at the bottom of the figure.  As we can see, PanGu-Draw generates high-quality images, which are better than or on par with these top-performing models.

